\documentclass[10pt,twocolumn,letterpaper]{article}

\usepackage[pagenumbers]{cvpr} % To force page numbers, e.g. for an arXiv version

\newcommand{\ours}{\texttt{SAB3R}~}
\usepackage{multirow} 
\usepackage{float} % Add this in your preamble

\definecolor{cvprblue}{rgb}{0.21,0.49,0.74}
\usepackage[pagebackref,breaklinks,colorlinks,allcolors=cvprblue]{hyperref}

\def\paperID{0000} % *** Enter the Paper ID here
\def\confName{CVPR}
\def\confYear{2025}

\title{SAB3R: Semantic-Augmented Backbone in 3D Reconstruction}

\author{Xuweiyi Chen$^{1}$\thanks{Equal contribution} \quad Tian Xia$^{2}$\footnotemark[1] \quad Sihan Xu$^2$ \quad Jianing Yang$^2$ \\
Joyce Chai$^{2}$ \quad Zezhou Cheng$^{1}$ \\
$^1$ University of Virginia \quad $^2$ University of Michigan\\
\url{https://uva-computer-vision-lab.github.io/sab3r/}
}

\begin{document}
\maketitle
\begin{abstract}

We introduce a new task, \textbf{Map and Locate}, which unifies the traditionally distinct objectives of open-vocabulary segmentation—detecting and segmenting object instances based on natural language queries—and 3D reconstruction, the process of estimating a scene’s 3D structure from visual inputs. Specifically, Map and Locate involves generating a point cloud from an unposed video and segmenting object instances based on open-vocabulary queries. This task serves as a critical step toward real-world embodied AI applications and introduces a practical task that bridges reconstruction, recognition and reorganization.

To tackle this task, we introduce a simple yet effective baseline, which we denote as \textbf{\ours}. Our approach builds upon MASt3R, a recent breakthrough in 3D computer vision, and incorporates a lightweight distillation strategy. This method transfers dense, per-pixel semantic features from 2D vision backbones (\eg, CLIP and DINOv2) to enhance MASt3R’s capabilities. Without introducing any auxiliary frozen networks, our model generates per-pixel semantic features and constructs cohesive point maps in a single forward pass. 

Compared to separately deploying MASt3R and CLIP, our unified model, \ours, achieves superior performance on the Map and Locate benchmark. Furthermore, we evaluate \ours on both 2D semantic segmentation and 3D tasks to comprehensively validate its effectiveness.
\vspace{-5px}
\end{abstract}
 
\section{Introduction}
\label{sec:intro}
% \makeatletter
% \let\@oldmaketitle\@maketitle
% \renewcommand{\@maketitle}{\@oldmaketitle
%     \centering
%     \includegraphics[width=0.6\linewidth]{figs/source/teaser_v4.pdf}
%     \captionof{figure}{Given an unposed input video (a), we show ground truth for: 
% (b) open-vocab semantic segmentation (per-pixel labels for the prompt “a black office chair”), 
% (c) 3D reconstruction (ground-truth point cloud), and 
% (d) the proposed \emph{Map and Locate} task (open-vocab segmentation for the prompt “a black office chair” and point cloud). 
% The Map and Locate task: (1) encompasses both 2D and 3D tasks, 
% (2) bridges reconstruction and recognition, 
% and (3) introduces practical questions in robotics and embodied AI. The
% Map and Locate generalizes both 2D and 3D tasks, and we expect this unified approach to present novel challenges and enable innovative new methods. \vspace{-5pt}}
%     \label{fig:teaser}
%     \bigskip
% }

\begin{figure}[t]
% \begin{wrapfigure}{r}{0.45\textwidth}
    \centering
    \includegraphics[width=0.8\linewidth]{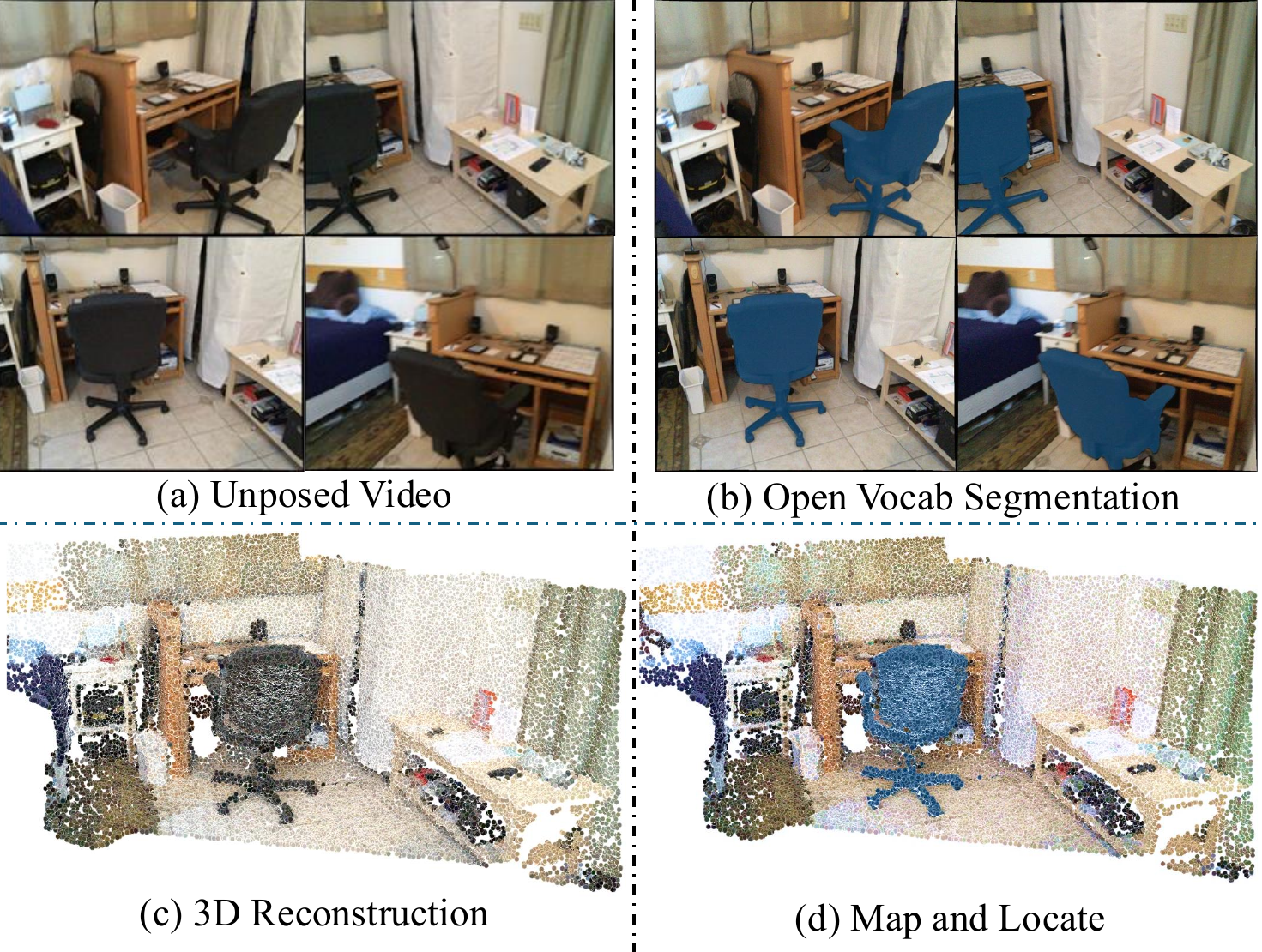}
\caption{Given an unposed input video (a), we show ground truth for: 
(b) open-vocab semantic segmentation (per-pixel labels for the prompt “a black office chair”), 
(c) 3D reconstruction (ground-truth point cloud), and 
(d) the proposed \emph{Map and Locate} task (open-vocab segmentation for the prompt “a black office chair” and point cloud). 
The Map and Locate task: (1) encompasses both 2D and 3D tasks, 
(2) bridges reconstruction and recognition, 
and (3) introduces practical questions in robotics and embodied AI. The
Map and Locate generalizes both 2D and 3D tasks, and we expect this unified approach to present novel challenges and enable innovative new methods.}
    \vspace{-10px}
    \label{fig:teaser}
% \end{wrapfigure}
\end{figure}

% \makeatletter
% \let\@oldmaketitle\@maketitle
% \renewcommand{\@maketitle}{\@oldmaketitle
%     \centering
%     \includegraphics[width=\linewidth]{figs/source/teaser_v4.pdf}
%     \vspace*{-20pt}
%     \captionof{figure}{Our method, semantic-augmented backbone for 3D reconstruction, dubbed \textbf{\ours}, enables zero-shot open-vocabulary segmentation and 3D reconstruction from unposed images in a single forward pass. Building on a state-of-the-art 3D foundation model, \ours generates dense 2D foundational features, such as CLIP~\cite{radford2021learningtransferablevisualmodels} and DINOv2~\cite{oquab2024dinov2learningrobustvisual}. Additionally, \ours introduces a new capability by simultaneously performing 3D reconstruction and open-vocabulary semantic segmentation.\vspace{-5px}}
%     \label{fig:teaser}
%     \bigskip
% }

\begin{figure*}[t!]
    \centering
    \includegraphics[width=0.9\linewidth]{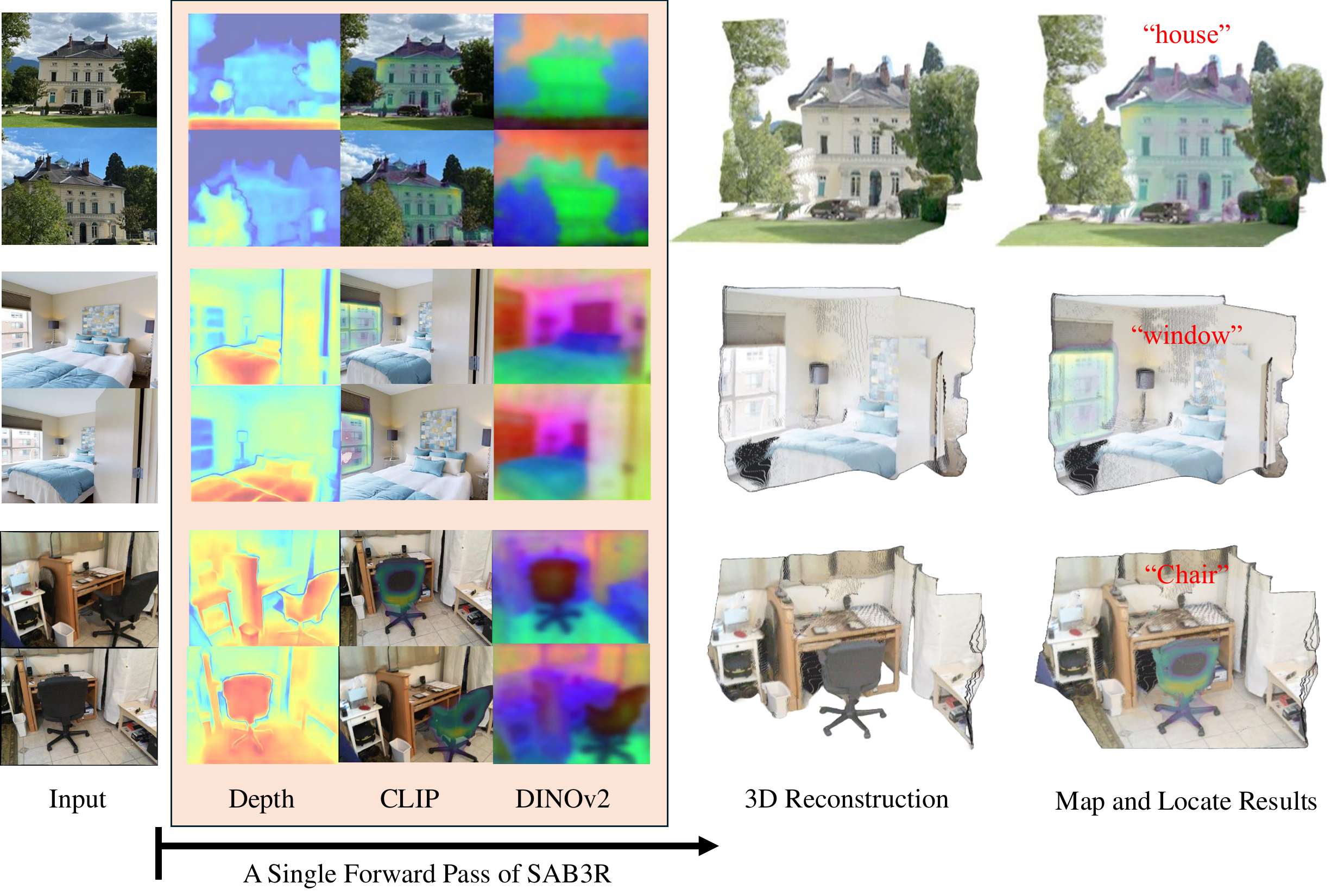}
    \vspace{-5px}
\caption{Our method, \ours, a semantic-augmented backbone for 3D reconstruction, enables zero-shot open-vocabulary segmentation and 3D reconstruction from unposed images in a single forward pass. By jointly performing reconstruction and open-vocabulary semantic segmentation, \ours~introduces a novel capability that unifies these tasks within a single framework. \vspace{-10px}}
    \label{fig:main_vis}
\end{figure*}

Current 3D open-vocabulary segmentation methods~\cite{Schult23ICRA,Peng2023OpenScene,yang2024llm} typically assume access to complete, high-quality point clouds—an assumption that rarely holds in real-world embodied AI scenarios. One major challenge lies in the high cost and complexity of curating large-scale 3D open-vocabulary datasets, even with prior efforts such as ScanRefer~\cite{chen2020scanrefer} and ReferIt3D~\cite{achlioptas2020referit3d}, which remain limited in both scale and diversity. Additionally, existing methods either depend on precise camera poses and sensor calibration for accurate point cloud reconstruction, an impractical requirement in continuously changing environments, or rely on test-time optimization techniques~\cite{mildenhall2020nerf, kerbl3Dgaussians}, which are computationally expensive and unsuitable for real-time applications. Despite these challenges, human perception effortlessly integrates visual semantics with 3D structural understanding, leveraging depth cues and object motion over a lifetime of interaction~\cite{6277403}. Thus, we aim to explore how a model can simultaneously perform semantic understanding and 3D reconstruction.

Malik et al.~\cite{MALIK20164} categorize vision tasks into recognition, reconstruction, and reorganization. Recognition involves assigning semantic categories to images, reconstruction focuses on estimating 3D structures, and reorganization deals with grouping and segmenting images based on spatial or perceptual similarity. Ideally, these tasks should mutually benefit one another. Moreover, maintaining separate models for different vision tasks is inefficient, incurring high memory and runtime costs~\cite{sanh2022multitaskpromptedtrainingenables}. This raises a critical question: \textit{Can 3D open-vocabulary segmentation and 3D reconstruction be effectively reconciled?}

To address this challenge, we propose a new approach that takes unposed video as input—a natural and accessible modality for embodied agents operating in real-world settings. Unlike existing methods that rely on posed RGB-D sequences or pre-scanned environments, our method requires neither pre-computed point clouds nor precise camera calibration. As illustrated in Fig.~\ref{fig:teaser}, we introduce the \textit{Map and Locate} task, which jointly constructs a 3D geometric map and segments objects specified through open-vocabulary queries. This formulation enables simultaneous spatial mapping, semantic understanding, and object localization from raw visual input. To this end, we present a simple yet effective baseline, \ours, shown in Fig.~\ref{fig:main_vis}, which takes unposed images as input and predicts a point map, dense CLIP features, and dense DINOv2 features in a single forward pass. We present qualitative results for 3D Reconstruction and \textit{Map and Locate} task in Fig.~\ref{fig:teaser}.

% Therefore, our work addresses this challenge by drawing inspiration from human visual perception. As human can seamlessly interpret images by combining 2D visual information with an intuitive understanding of 3D structure. While existing methods take in posed RGB-D sequences or pre-scanned environments, we propose using unposed video as input—a natural and accessible modality for embodied agents operating in real-world settings. As illustrated in Fig.~\ref{fig:teaser}, our Map and Locate task jointly constructs a 3D geometric map and segments objects specified through open-vocabulary queries. This approach enables simultaneous spatial mapping, semantic understanding, and segmentation without requiring pre-processed point clouds. To this end, we introduce a simple yet effective baseline \ours, as shown in Fig.~\ref{fig:main_vis}, which takes unposed images as input and predicts a point map, dense CLIP features, and dense DINOv2 features in a single forward pass. 

This integration offers three key advantages. First, it eliminates the reliance on high-quality, pre-scanned point clouds by taking in unposed video as input. Second, it removes the dependence on precise camera poses and sensor calibrations, making 3D segmentation and reconstruction feasible in real-world environments without test-time optimization, which is often computationally prohibitive. Third, it unifies recognition, reorganization and reconstruction into a single model, reducing memory and runtime overhead. By bridging the gap between open-vocab segmentation and reconstruction, our approach offers a more practical and scalable solution for embodied perception. 

In summary, our contributions are:
\begin{itemize}
    \item \textit{Map and Locate} Benchmark: We introduce a novel benchmark for multi-view 3D semantic segmentation that jointly addresses the tasks of reconstruction, reorganization, and recognition. The benchmark is accompanied by a large-scale dataset, clearly defined evaluation protocols, and standardized metrics.
    \item \ours: We propose a unified framework that concurrently performs open-vocabulary segmentation and 3D reconstruction from unposed images via an efficient distillation strategy. We present it as a baseline due to its performance and computational efficiency.
\end{itemize}
\section{Related Work}
\label{sec:related}

\subsection{3D Reconstruction}

The landscape of 3D reconstruction has evolved from traditional geometric methods like SfM~\cite{agarwal2011building,schoenberger2016sfm} and SLAM~\cite{Cadena_2016,mur2015orb} to learning-based approaches that leverage data-driven priors~\cite{teed2021droid,wang2024vggsfm}. DUSt3R~\cite{dust3r} pioneered a paradigm shift by predicting dense point maps from image pairs in a shared coordinate frame, removing the need for explicit pose supervision. However, its reliance on stereo inputs limits its applicability to multi-view settings. More recently, MASt3R~\cite{mast3r_arxiv24} extended this idea by learning viewpoint-invariant representations for dense point prediction across multiple images, significantly improving robustness in unposed scenarios. While these advances enable reconstructing 3D geometry from unconstrained image sequences, they primarily focus on geometric consistency and do not incorporate high-level semantics.

Our work builds upon MASt3R and extends it to the novel Map and Locate task, which bridges 3D reconstruction with open-vocabulary segmentation. Unlike prior methods that treat reconstruction and recognition as separate problems, we introduce a unified approach that simultaneously maps the environment and segments objects based on free-form queries. This perspective transforms 3D perception into a richer and more interactive task, opening new avenues for embodied AI and scene understanding beyond purely geometric reconstruction.

\subsection{Leveraging 2D for 3D Vision}

Most 3D visual-language models operate directly on 3D point clouds without leveraging 2D pre-trained features. SAT-2D~\cite{yang2021sat2dsemanticsassisted} was one of the first 3D visual grounding models to incorporate 2D visual features, aligning 2D and 3D representations during training and achieving significant improvements over versions without 2D features. More recent approaches, such as 3DLLM~\cite{hong20233dllminjecting3dworld} in 3D Question Answering, use multi-view 2D features with LLMs to decode answers, but have yet to fully address 3D visual grounding tasks. Similarly, PQ3D~\cite{zhu2024unifying} integrates various visual backbones, including a 2D feature backbone from OpenScene~\cite{Peng2023OpenScene}.

EFM3D~\cite{straub2024efm3dbenchmarkmeasuringprogress} lifts 2D image features into 3D feature volumes, but focuses on 3D object detection and surface reconstruction. ODIN~\cite{jain2024odinsinglemodel2d} proposes an interleaved 2D-3D backbone with pre-trained 2D weights, but is limited to object detection. Fit3D~\cite{yue2024improving2dfeaturerepresentations}, which lifts 2D semantic features into 3D Gaussian representations, injects 3D awareness when training 2D foundation models—a complementary approach to ours.

\subsection{3D Open-Vocabulary Segmentation}

Our work is closely related to recent efforts in distilling 2D semantic features into 3D representations for open-vocabulary segmentation. These approaches often utilize neural rendering techniques, such as NeRF~\cite{mildenhall2020nerf} and Gaussian Splatting~\cite{kerbl3Dgaussians}, to aggregate multi-view information. For instance, Semantic NeRF~\cite{Zhi:etal:ICCV2021} and Panoptic Lifting~\cite{Siddiqui_2023_CVPR} embed 2D semantics into 3D volumes, enabling dense scene understanding.

More recent works, such as LeRF~\cite{kerr2023lerflanguageembeddedradiance}, Distilled Feature Fields~\cite{shen2023distilledfeaturefieldsenable}, NeRF-SOS~\cite{fan2022nerfsosanyviewselfsupervisedobject}, and Neural Feature Fusion Fields~\cite{tschernezki2022neuralfeaturefusionfields}, further distill features from strong 2D models like LSeg~\cite{li2022languagedrivensemanticsegmentation} and DINO~\cite{caron2021emergingpropertiesselfsupervisedvision} into view-consistent 3D representations. Featured 3DGS~\cite{zhou2024feature} extends this paradigm to the Gaussian Splatting framework, enabling efficient distillation of 2D pre-trained models into 3D point-based representations.

While prior methods have demonstrated strong performance in 3D open-vocabulary segmentation, they typically depend on posed multi-view images and scene-specific optimization, which constrains their applicability in real-world settings. In contrast, our approach eliminates the need for pose supervision by directly distilling 2D features into point maps, enabling broader generalization across diverse and unstructured environments.  

Similarly, LSM~\cite{lsm} jointly estimates geometry, appearance, and semantics in a single feed-forward pass and is capable of synthesizing diverse label maps. However, it employs a frozen language segmentation backbone and restricts input to only two images due to its reliance on point transformer~\cite{wu2024ptv3}.
\section{A Novel Task: \textit{\textit{Map and Locate}}}

\paragraph{Task Setting}

In this novel task, termed \textit{Map and Locate}, the model receives multiview inputs and a set of semantic labels to reconstruct a 3D scene and localize target objects based on text prompts. This task extends beyond independent depth estimation for each image, requiring the model to infer relative camera poses across views and classify the semantic category of each predicted 3D point.

The task is defined as follows: given \(n\) input images (\(n \geq 2\)) and a set of grounding queries \(\mathcal{L} = \{0, \dots, L-1\}\), the goal is to map each pixel \(i\) to a pair \((X_i, l_i) \in \mathbb{R}^3 \times \mathcal{L}\), where \(X_i = (x_i, y_i, z_i)\) represents the 3D coordinates of the point corresponding to pixel \(i\), and \(l_i\) denotes its semantic class. For an image \(I\) of resolution \(W \times H\), this establishes a one-to-one mapping between pixels and 3D scene points with semantic labels, i.e., \(I_{i,j} \leftrightarrow (X_{i,j}, l_{i,j})\), for all \((i, j) \in \{1, \dots, W\} \times \{1, \dots, H\}\). We assume each camera ray intersects only a single 3D point, excluding cases like translucent surfaces. Ambiguous or out-of-class pixels are assigned a void label in the annotations.

For implementation, we adopt MaskCLIP~\cite{dong2023maskclipmaskedselfdistillationadvances} enhanced with FeatUp~\cite{fu2024featup}, combined with the MASt3R~\cite{mast3r_arxiv24} pipeline as our baseline method. MaskCLIP and MASt3R act as teacher models for \ours, guiding the distillation process to achieve both 3D reconstruction and open-vocabulary semantic segmentation.

\paragraph{Data Curation}

Our data is sourced from ScanNet~\cite{dai2017scannet}, a large-scale indoor scene dataset that provides RGB-D sequences, camera poses, and semantic and instance-level annotations. From the validation split, we curate a subset of 24 diverse scenes, selected based on their unique object layouts and camera trajectories. For each scene, we create image groups containing 2, 3, or 4 views, ensuring that each image overlaps with at least one other in the group. This overlap guarantees shared visual context, enabling robust evaluation of 3D reconstruction and localization tasks. To balance evaluation time and dataset diversity, we limit our selection to 24 scenes, which already requires approximately 10 hours for the evaluation to complete.

For semantic classification, we map ground-truth annotations to the widely used NYU40 class taxonomy~\cite{Silberman:ECCV12}. The curated dataset includes a wide range of objects with both semantic and instance-level annotations. Each image group is paired with its corresponding RGB images, depth maps, camera poses (intrinsics and extrinsics), and semantic and instance labels. Detailed data statistics, example image groups, and the full data curation process, including selection criteria and preprocessing steps, are provided in the supplementary materials.

\vspace{-10pt}
\paragraph{Evaluation Metrics}

For the \textit{Map and Locate} task, we evaluate model performance using several key metrics, and in all metrics, higher values consistently indicate better performance. 
Additionally, before evaluating these metrics, models are required to compute pair $(X, l)$ for every pixel in each image, using only the image inputs without any ground truth data, such as intrinsic or extrinsic matrices, then use one ground truth image’s depth and pose for scaling and alignment to the ground truth coordinates.

\textit{mIoU} (mean intersection over union) quantifies the overlap between predicted and ground truth points, calculated as the ratio of correctly predicted points to the union of predicted and ground truth points. This metric provides an overall measure of segmentation accuracy. In our task, we compute the mIoU by finding the nearest predicted point for each ground truth point and using its label to evaluate against the ground truth labels.

\textit{Acc} (accuracy) is defined as the proportion of correctly predicted points relative to the total ground truth points, indicating the model’s effectiveness in assigning correct semantic classes to 3D points. In our setting, similar to mIoU, we calculate Acc using the same approach.

\textit{mComp} (Mean Completeness) measures how comprehensively the predicted points cover the ground truth point cloud. After aligning the predicted points with the ground truth pose, we compute the average distance from each predicted point to its nearest neighbor in the ground truth, offering a general sense of the reconstruction's completeness. For our task, we filter points based on each test label in both the ground truth and the predictions, then calculate the mComp metric accordingly.
 
\textit{mdComp} (Median Completeness) is similar to mean completeness but calculates the median of nearest-neighbor distances instead. This approach reduces the impact of outliers, providing a more stable indication of coverage consistency across samples.

\section{Method}
\label{sec:method}
\begin{figure*}[t!]
    \centering
    \includegraphics[width=0.9\linewidth]{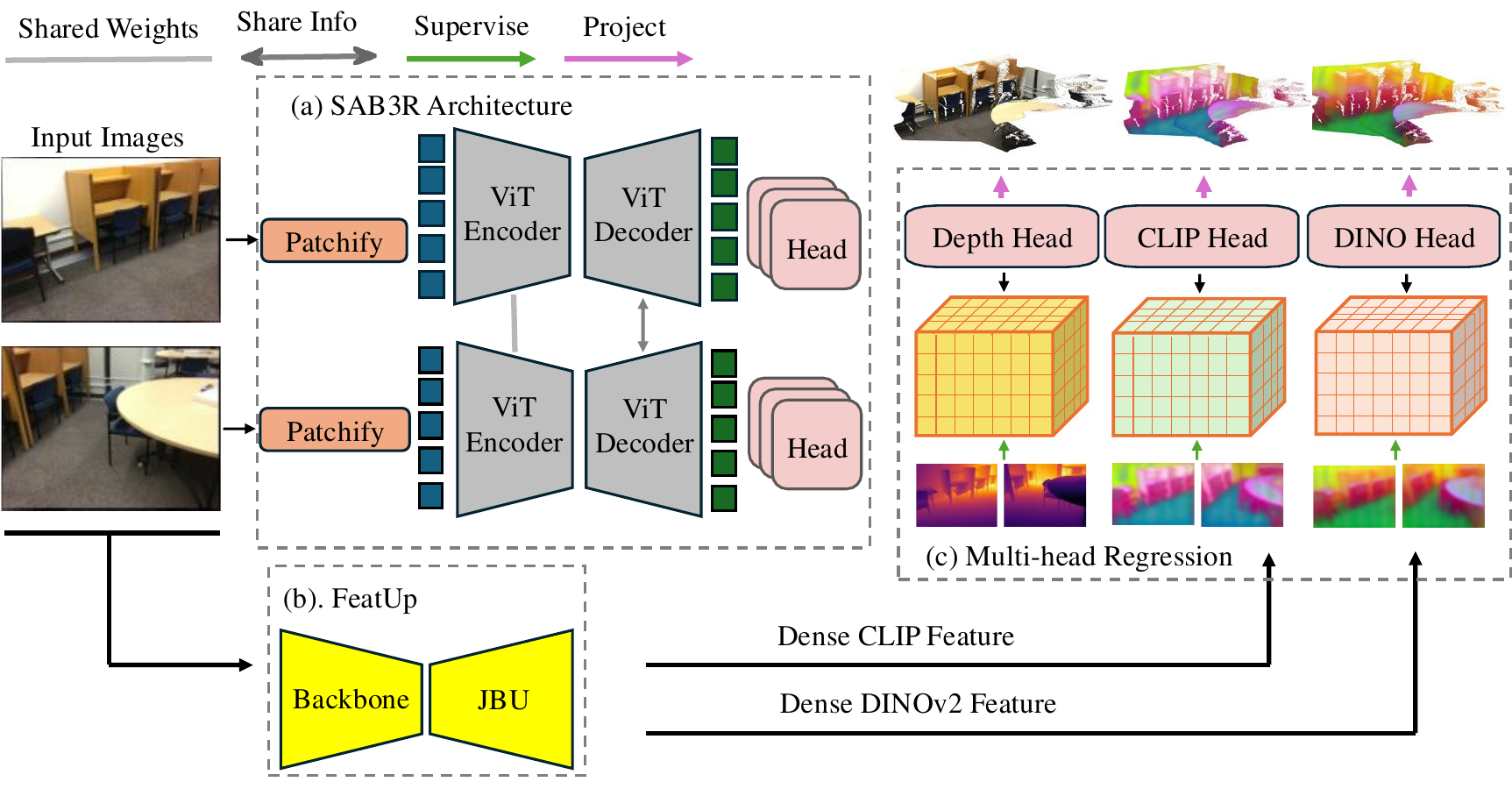}
    \caption{\textbf{Methods Architecture.} We distill dense features from CLIP and DINO into the MASt3R framework, enriching it with 2D semantic understanding. Each encoder-decoder pair operates on multi-view images, sharing weights and exchanging information to ensure consistent feature extraction across views. The model simultaneously generates depth, dense DINOv2, and dense CLIP features, which are then used for multi-view 3D reconstruction and semantic segmentation. This architecture enables \ours to seamlessly integrate 2D and 3D representations, achieving both geometric and semantic comprehension in a unified model. \vspace{-10pt}}
    
    \label{fig:main}
\end{figure*}

In this section, we present \ours, a simple baseline method that distills dense 2D semantic features from foundation models into a 3D reconstruction framework. Building on a base 3D reconstruction model, we transfer knowledge from 2D foundation features—enhanced via FeatUp~\cite{fu2024featup}—to integrate semantic understanding into the 3D domain. Our objective is to unify 2D and 3D representations within a shared backbone, enabling joint 3D reconstruction and open-vocabulary semantic segmentation.

To facilitate understanding, this section is organized as follows: Sec.~\ref{subsec:prior_dust3r} reviews the core 3D reconstruction backbone, Sec.~\ref{subsec:2d_distillation} details the distillation process of 2D semantic features, and Sec.~\ref{subsec:feature_extension} outlines how additional features can be incorporated to further enrich the model’s capabilities.

\subsection{Foundational Components}
\label{subsec:prior_dust3r}
DUSt3R~\cite{dust3r} is a recent method that addresses a range of 3D tasks using unposed images as input, including camera calibration, depth estimation, pixel correspondence, camera pose estimation, and dense 3D reconstruction. It uses a transformer-based network to generate \emph{local} 3D reconstructions from two input images, producing dense 3D point clouds \( X^{1, 1} \) and \( X^{2, 1} \), referred to as \emph{pointmaps}.

A pointmap \( X^{a, b} \in \mathbb{R}^{H \times W \times 3} \) represents a 2D-to-3D mapping from each pixel \( i = (u, v) \) in image \( I^{a} \) to its corresponding 3D point \( X^{a, b}_{u, v} \in \mathbb{R}^3 \) in the coordinate system of camera \( C^{b} \). By jointly regressing two pointmaps, \( X^{1, 1} \) and \( X^{2, 1} \), expressed in the coordinate system of camera \( C^{1} \), DUSt3R simultaneously performs calibration and 3D reconstruction. For multiple images, a global alignment step merges all pointmaps into a unified coordinate system.

Images are encoded in a Siamese manner using a ViT~\cite{vit}, producing representations \( H^1 \) and \( H^2 \):
\[
    H^1 = \text{Encoder}(I^1), \quad H^2 = \text{Encoder}(I^2).
\]
Two intertwined decoders process these representations, exchanging information via cross-attention to capture spatial relationships and global 3D geometry. The enhanced representations are denoted \( H'^1 \) and \( H'^2 \):
\[
    H'^1, H'^2 = \text{Decoder}(H^1, H^2).
\]
Finally, prediction heads regress the pointmaps and confidence maps:
\begin{align}
    X^{1, 1}, C^1 &= \text{Head}^1_{3D}([H^1, H'^1]), \\
    X^{2, 1}, C^2 &= \text{Head}^2_{3D}([H^2, H'^2]).
\end{align}

\subsection{Distilling 2D Semantic Features}
\label{subsec:2d_distillation}
To integrate 2D semantic information into the model while retaining its 3D capabilities, we design a multitask framework that prevents catastrophic forgetting. This framework enables the model to simultaneously learn both 2D and 3D features. We adopt the MASt3R~\cite{mast3r_arxiv24} architecture, which consists of a ViT-Large encoder, a ViT-Base decoder, and DPT heads. To distill dense 2D features, we introduce new heads to regress features from DINO~\cite{oquab2024dinov2learningrobustvisual} and CLIP~\cite{radford2021learningtransferablevisualmodels}. 

Following DUSt3R~\cite{dust3r} and MASt3R~\cite{leroy2024groundingimagematching3d}, the new heads leverage either a DPT architecture or a simpler MLP structure. The DPT design is particularly effective for dense prediction tasks like depth estimation and semantic feature extraction. In addition to the depth and descriptor heads (\(\text{Head}_{3D}^{1, 2}\) and \(\text{Head}_{desc}^{1, 2}\)), we introduce two new heads, \(\text{Head}_{\text{2D feature}}^{1, 2}\), for distilling 2D features:
\begin{align}
    S^{1} &= \text{Head}^1_{\text{2D feature}}([H^1, H'^1]), \\
    S^{2} &= \text{Head}^2_{\text{2D feature}}([H^2, H'^2]).
\end{align}
Here, \( H^1 \) and \( H^2 \) are embeddings from the encoder, and \( H'^1 \), \( H'^2 \) are enhanced representations from the decoder. The concatenation \([H, H']\) combines multi-scale features from each view.

To preserve depth estimation capabilities, we retain the regression loss \(\mathcal{L}_{conf}\) from DUSt3R and the matching loss \(\mathcal{L}_{match}\) from MASt3R. Additionally, we introduce a regression loss for the 2D features, guiding the model to learn semantic information:
\begin{equation}
    \mathcal{L}_{\text{2D}} = \left\Vert S^{v} - \hat S^{v} \right\Vert, \quad v \in \{1, 2\},
\end{equation}
where \(\hat S^{v}\) is the target 2D feature extracted from foundation models for the corresponding view \(v\). Dense pixel features from FeatUp~\cite{fu2024featup} are used as supervision.

The total loss combines all components, weighted by hyper-parameters \(\beta\) and \(\gamma\):
\begin{equation}
    \mathcal{L}_{\text{total}} = \mathcal{L}_{conf} + \beta \mathcal{L}_{match} + \gamma \mathcal{L}_{\text{2D}}.
\end{equation}

\subsection{Incorporating Additional Features}
\label{subsec:feature_extension}
Our distillation pipeline is designed to flexibly incorporate multiple 2D features into the 3D foundation model, enhancing its capabilities. For each additional feature, we add a dedicated head and regression loss, resulting in an updated training objective:
\begin{equation}
    \mathcal{L}_{\text{total}} = \mathcal{L}_{conf} + \beta \mathcal{L}_{match} + \gamma_1 \mathcal{L}_{2D_1} + \gamma_2 \mathcal{L}_{2D_2}.
\end{equation}
Here, \(\mathcal{L}_{2D_1}\) and \(\mathcal{L}_{2D_2}\) are regression losses for individual 2D features, with \(\gamma_1\) and \(\gamma_2\) controlling their contributions. MaskCLIP and DINOv2 features are integrated into the 3D backbone through this framework, with dedicated heads for each feature.
\section{Experiments}
\label{sec:exp}

In this section, we showcase the effectiveness of our simple baseline \ours for distilling 2D foundation models into a 3D reconstruction model. The section is organized into five parts. In Sec.~\ref{sec:implemen_detail}, we provide details of our implementation for \ours. Sec.~\ref{subsec:3d} analyzes how \ours retains 3D performance compared to the teacher models. In Sec.~\ref{subsec:open-vocab}, we demonstrate our method’s zero-shot semantic segmentation performance, achieving results comparable to the teacher models. Finally, in Sec.~\ref{sec:map_and_locate}, we present results and analysis for the novel task, \textit{Map and Locate}. 

\subsection{Implementation Details}
\label{sec:implemen_detail}

We fine-tune our model based on pre-trained MASt3R~\cite{mast3r_arxiv24} with datasets from DUSt3R~\cite{dust3r} and MASt3R~\cite{mast3r_arxiv24}, including Habitat~\cite{szot2021habitat}, ScanNet++~\cite{yeshwanth2023scannethighfidelitydataset3d}, ARKitScenes~\cite{dehghan2021arkitscenes}, Co3Dv2~\cite{reizenstein21co3d}, and BlenderMVS~\cite{yao2020blendedmvs}. Data preprocessing adheres to the guidelines of each dataset. To avoid the impracticality of storing dense 2D VFM features locally, which would require over 60 TB of storage, we leverage FeatUp to dynamically generate these features during training. Additional details on the datasets and preprocessing steps are provided in the supplementary materials.

\vspace{-10pt}
\paragraph{Training} We adopt MASt3R~\cite{mast3r_arxiv24} as the base 3D foundation model. During training, we unfreeze the encoder to improve its ability to extract semantically meaningful 2D features while preserving depth estimation accuracy. For distillation using only MaskCLIP features, we set the loss weights to $\beta = 0.75$ and $\gamma = 20$. When distilling both MaskCLIP and DINOv2 features, we modify the weights to $\beta = 0.75$, $\gamma_1 = 20$, and $\gamma_2 = 4$. Based on our empirical observations, these hyperparameters are highly sensitive—small deviations can result in modality collapse.

% \vspace{-10pt}
% \paragraph{Computational Resources} Each checkpoint is optimized around 3 days, using either 8 A40 GPUs or 4 A100 80GB GPUs.

\subsection{Zero-Shot 3D Tasks}
\label{subsec:3d}
\begin{table}[t!]
\begin{center}
\small
\renewcommand\arraystretch{1.1}
\hspace{-3mm}
\resizebox{0.45\textwidth}{!}{
\begin{tabular}{lcccccc}
% \hline
\specialrule{1.5pt}{0.5pt}{0.5pt} 
Methods & Train & \multicolumn{2}{c}{NYUD-v2 (Indoor)} & \multicolumn{2}{c}{KITTI (Outdoor)} \\
\cline{3-6}
 &  & Rel$\downarrow$ & $\delta_{1.25}\uparrow$ & Rel$\downarrow$ & $\delta_{1.25}\uparrow$ \\
 \hline
% \specialrule{1.5pt}{0.5pt}{0.5pt} 
DPT-BEiT\cite{dpt}                     &D& {\bf 5.40}&{\bf 96.54} & 9.45 & 89.27 \\
NeWCRFs\cite{yuan2022newcrfsneuralwindow}   &D& 6.22&95.58 & {\bf 5.43} & {\bf 91.54} \\
\hline
Monodepth2~\cite{monodepth2}      &SS& 16.19&74.50 & 11.42 & 86.90 \\
SC-SfM-Learners~\cite{bian2021ijcv} &SS& 13.79&79.57 & 11.83 & 86.61 \\
SC-DepthV3~\cite{sc_depthv3}     &SS&{\bf 12.34}&{\bf 84.80} & 11.79 & 86.39 \\
MonoViT~\cite{monovit}  &SS& - & - &  {\bf 9.92} & {\bf 90.01} \\
\hline
RobustMIX~\cite{ngnawe2024robustmiximprovingrobustnessregularizing}   &T& 11.77 & 90.45 & 18.25 & 76.95 \\
SlowTv~\cite{spencer2023kickrelaxlearning}        &T& 11.59&87.23 & (6.84) & (56.17) \\ %
{DUSt3R 224-NoCroCo}        &T& 14.51 & 81.06 & 20.10 & 71.21 \\ 
{DUSt3R 224}        &T& 10.28&88.92 & 16.97 & 77.89 \\ 
{DUSt3R 512}    &T&  \textbf{6.51} & \textbf{94.09} & \textbf{12.02} &\textbf{83.43} \\ %
\hline{
MASt3R}    &T& 8.17 &92.59 & \textbf{8.28} & \textbf{93.27} \\ %
\ours (C)   &T& 7.80 & 92.67 & 11.63 & 86.74 \\ %
\ours (CD)    &T& \textbf{7.67} & \textbf{92.82} & 12.53 & 83.51 \\ %
\specialrule{1.5pt}{0.5pt}{0.5pt} 
\end{tabular}}
\normalsize
\end{center}
% \vspace{-10pt}
\caption{\textbf{Monocular depth estimation on NYU-v2 and KITTI datasets.} 
D = Supervised, SS = Self-supervised, T = Transfer (zero-shot). (Parentheses) refers to training on the same set. \ours(C) represents our model distilled with CLIP features, while \ours(CD) builds upon this by integrating both CLIP and DINO features during distillation. This notation is used consistently throughout the paper.
\label{tab:monocular-depth}
}
\end{table}
\begin{table}[t!]
\begin{center}
\small
\renewcommand\arraystretch{1.1}
\hspace{-3mm}
\resizebox{0.45\textwidth}{!}{
\begin{tabular}{lccc}
% \hline
\specialrule{1.5pt}{0.5pt}{0.5pt} 
Methods & RRA@15$\uparrow$ & RTA@15$\uparrow$ & mAA(30)$\uparrow$ \\ 
% \specialrule{1.5pt}{0.5pt}{0.5pt} 
\hline
Colmap+SG~\cite{superpoint, superglue} & 36.1 & 27.3 & 25.3 \\
PixSfM~\cite{pixsfm}                  & 33.7 & 32.9 & 30.1 \\
RelPose~\cite{relpose}                & 57.1 & -    & -    \\
PosReg~\cite{posediffusion}           & 53.2 & 49.1 & 45.0 \\
PoseDiff~\cite{posediffusion}         & 80.5 & 79.8 & 66.5 \\
RelPose++~\cite{relposepp}            & (85.5) & -   & -    \\
RayDiff~\cite{raydiffusion}           & (93.3) & -   & -    \\ 
DUSt3R-GA~\cite{dust3r}               & \textbf{96.2} & 86.8 & 76.7 \\
\hline
DUSt3R~\cite{dust3r}                  & 94.3 & 88.4 & 77.2 \\
MASt3R                              & 94.2 & \textbf{88.6} & \textbf{81.1} \\ 
\ours (C)                           & 92.6 & 87.3 & 79.7 \\ 
\ours (CD)                          & 92.9 & 87.8 & 80.3 \\ 
\specialrule{1.5pt}{0.5pt}{0.5pt} 
\end{tabular}}
\normalsize
\end{center}
% \vspace{-10pt}
\caption{\textbf{Multi-view pose regression on the CO3Dv2~\cite{reizenstein21co3d} dataset using 10 randomly selected frames.} For methods that do not report results for the 10-view setup, we include their 8-view performance in parentheses. We distinguish between multi-view and pairwise methods for clarity. Notably, our method performs competitively with state-of-the-art approaches.}

\label{tab:multi-view-pose-regression}
\end{table}
\begin{table*}[t!]
\begin{center}
\resizebox{\linewidth}{!}{
\begin{tabular}{lcc|cccccccccccc}
\toprule
Model & \textbf{Params} & \textbf{FLOPs} 
& \multicolumn{4}{c}{Sparse View = 2} 
& \multicolumn{4}{c}{Sparse View = 3} 
& \multicolumn{4}{c}{Sparse View = 4} \\
\cmidrule(lr){4-7} \cmidrule(lr){8-11} \cmidrule(lr){12-15}
& & 
& \textbf{mIoU} & \textbf{Acc.} & \textbf{mComp.} & \textbf{mdComp.} 
& \textbf{mIoU} & \textbf{Acc.} & \textbf{mComp.} & \textbf{mdComp.}  
& \textbf{mIoU} & \textbf{Acc.} & \textbf{mComp.} & \textbf{mdComp.}  \\
\midrule
Baseline                 
& 838M & 248G 
& 4.57 & 18.10 & 0.64 & 0.67 
& 6.03 & 21.26 & 0.68 & 0.71 
& 5.12 & 19.31 & 0.68 & 0.70 \\
LSM~\cite{lsm}
& 1B & $\textgreater$ 592G 
& \textbf{21.40} & 42.34 & 0.72 & \textbf{0.80}  
& -  & - & - & - 
& - & - & - & - \\
\ours (C)               
& 729M & 218G
& 17.26 & 41.11 & \textbf{0.73} & 0.75 
& 22.83 & \textbf{53.19} & \textbf{0.78} & \textbf{0.81} 
& 19.92 & \textbf{48.07} & \textbf{0.77} & \textbf{0.80} \\
\ours (CD)              
& 729M & 218G 
& 17.50 & \textbf{42.72} & \textbf{0.73} & 0.76
& \textbf{22.94} & 52.86 & 0.77 & 0.80 
& \textbf{20.31} & 46.26 & 0.75 & 0.78 \\
\bottomrule
\end{tabular}
}
\vspace{-15px}
\end{center}
\caption{Performance comparison across different sparse view configurations (2, 3, and 4 views) using mIoU, Accuracy, Mean Completeness, and Median Completeness. Params and FLOPs refer to the number of parameters and computational cost per frame.}
\label{tab:sparse_view_metrics}
\end{table*}

% \begin{table*}[t!]
% \begin{center}
% \resizebox{\linewidth}{!}{
% \begin{tabular}{lcccccccccccc}
% \toprule
% Model  & \multicolumn{4}{c}{Sparse View = 2} & \multicolumn{4}{c}{Sparse View = 3} & \multicolumn{4}{c}{Sparse View = 4} \\
% \cmidrule(lr){2-5} \cmidrule(lr){6-9} \cmidrule(lr){10-13}
% & \textbf{mIoU} & \textbf{Acc.} & \textbf{Comp.} & \textbf{Median Comp.} 
% & \textbf{mIoU} & \textbf{Acc.} & \textbf{Comp.} & \textbf{Median Comp.}  
% & \textbf{mIoU} & \textbf{Acc.} & \textbf{Comp.} & \textbf{Median Comp.}  \\
% \midrule
% Baseline                 & 4.57 & 18.10 & 0.64 & 0.67 &
% 6.03 & 21.26 & 0.68 & 0.71 &
% 5.12 & 19.31 & 0.68 & 0.70 \\
% \ours (C)               & 17.26 & 41.11 & \textbf{0.73} & 0.75 &
% 22.83 & \textbf{53.19} & \textbf{0.78} & \textbf{0.81} &
% 19.92 & \textbf{48.07} & \textbf{0.77} & \textbf{0.80} \\
% \ours (CD)              & \textbf{17.50} & \textbf{42.72} & \textbf{0.73} & \textbf{0.76} &
% \textbf{22.94} & 52.86 & 0.77 & 0.80 &
% \textbf{20.31} & 46.26 & 0.75 & 0.78 \\
% \bottomrule
% \end{tabular}
% }
% \vspace{-1.0em}
% \end{center}
% \caption{Performance comparison across different sparse view configurations (2, 3, and 4 views) using mIoU, Accuracy, Mean Completeness, and Median Completeness.}
% \label{tab:sparse_view_metrics}
% \end{table*}
\begin{table}[t]
\centering
\small
\resizebox{0.95\linewidth}{!}{
\begin{tabular}{lccc}
\specialrule{1.5pt}{0.5pt}{0.5pt}
Model & Arch & VOC$\uparrow$ & ADE20k$\uparrow$ \\ 
\hline
GroupViT~\citep{GroupViT}         & ViT-S & 52.3 & - \\
ViewCo~\citep{ren2023viewco}      & ViT-S & 52.4 & - \\
ViL-Seg~\citep{vil-seg}           & ViT-B & 37.3 & - \\
OVS~\citep{OVS}                   & ViT-B & 53.8 & - \\
CLIPpy~\citep{PERCEPTUAL}         & ViT-B & 52.2 & 13.5 \\
TCL~\citep{TCL}                   & ViT-B & 51.2 & 14.9 \\
SegCLIP~\citep{seg-clip}          & ViT-B & 52.6 & 8.7 \\
SAM-CLIP~\citep{wang2024samclipmergingvisionfoundation} & ViT-B & \textbf{60.6} & 17.1 \\ 
\hline
FeatUp (MaskCLIP)                 & -     & 51.2 & 14.3 \\
\ours (C)                         & ViT-B & 55.4 & 18.3 \\
\ours (CD)                        & ViT-B & 56.4 & \textbf{19.0} \\ 
\specialrule{1.5pt}{0.5pt}{0.5pt}
\end{tabular}
}
\vspace{2pt}
\captionof{table}{\textbf{Zero-shot Semantic Segmentation Comparison.} Performance comparison of zero-shot semantic segmentation with recent state-of-the-art methods. \textbf{Note:} Results for \ours are based solely on the CLIP-head output. \vspace{-5px}}
\label{tab:zeroshot_combined}
\end{table}
% % \begin{table}[ht]
% % \begin{center}
% % \hspace{-5mm}
% % % \small
% % \renewcommand\arraystretch{1.1}
% \resizebox{\linewidth}{!}{
%     \begin{tabular}{lcccc}
%     % \hline
%     \specialrule{1.5pt}{0.5pt}{0.5pt}
%     Model & Arch & VOC$\uparrow$ & ADE20k$\uparrow$ \\ 
%     % \specialrule{1.5pt}{0.5pt}{0.5pt}
%     \hline
%     GroupViT~\citep{GroupViT}         & ViT-S & 52.3 & - \\
%     ViewCo~\citep{ren2023viewco}      & ViT-S & 52.4 & - \\
%     ViL-Seg~\citep{vil-seg}           & ViT-B & 37.3 & - \\
%     OVS~\citep{OVS}                   & ViT-B & 53.8 & - \\
%     CLIPpy~\citep{PERCEPTUAL}         & ViT-B & 52.2 & 13.5 \\
%     TCL~\citep{TCL}                   & ViT-B & 51.2 & 14.9 \\
%     SegCLIP~\citep{seg-clip}          & ViT-B & 52.6 & 8.7 \\
%     SAM-CLIP~\cite{wang2024samclipmergingvisionfoundation} & ViT-B & \textbf{60.6} & 17.1 \\ 
%     \hline
%     FeatUp (MaskCLIP)                 & -     & 51.2 & 14.3 \\
%     \ours (C)                         & ViT-B & 55.4 & 18.3 \\
%     \ours (CD)                        & ViT-B & 56.4 & \textbf{19.0} \\ 
%     \specialrule{1.5pt}{0.5pt}{0.5pt}
%     \end{tabular}
%     }
% % \normalsize
% % \end{center}
% % \vspace{-10pt}
% \caption{\textbf{Zero-shot Semantic Segmentation Comparison.} Performance comparison of zero-shot semantic segmentation with recent state-of-the-art methods. \textbf{Note:} Results for \ours are based solely on the CLIP-head output.}
% \label{tab:zeroshot_combined}
% % \end{table}
\paragraph{Monocular Depth Estimation}

We benchmark \ours on both an indoor dataset, NYUv2~\cite{Silberman:ECCV12}, and an outdoor dataset, KITTI~\cite{Geiger2013IJRR}, comparing its performance to state-of-the-art methods in Tab.~\ref{tab:monocular-depth}. For monocular depth evaluation, we use two commonly applied metrics following DUSt3R~\cite{dust3r} and recent studies~\cite{bian2021autorectifynetworkunsupervisedindoor, spencer2023kickrelaxlearning}.

As shown in Tab.~\ref{tab:monocular-depth}, \ours demonstrates strong adaptability to both indoor and outdoor environments. Distilling dense features from MaskCLIP or DINOv2 into the MASt3R backbone does not degrade the model’s performance or induce catastrophic forgetting for indoor setting. Therefore, \ours is still capable of making accurate depth prediction. Interestingly, \ours trained with MaskCLIP, or with both MaskCLIP and DINOv2, outperforms the base model MASt3R on the NYUv2 indoor dataset~\cite{Silberman:ECCV12}. However, our approach performs less effectively in outdoor scenarios, likely due to the indoor-focused nature of our training data.
\begin{figure*}[t!]
    \centering
    \includegraphics[width=\linewidth]{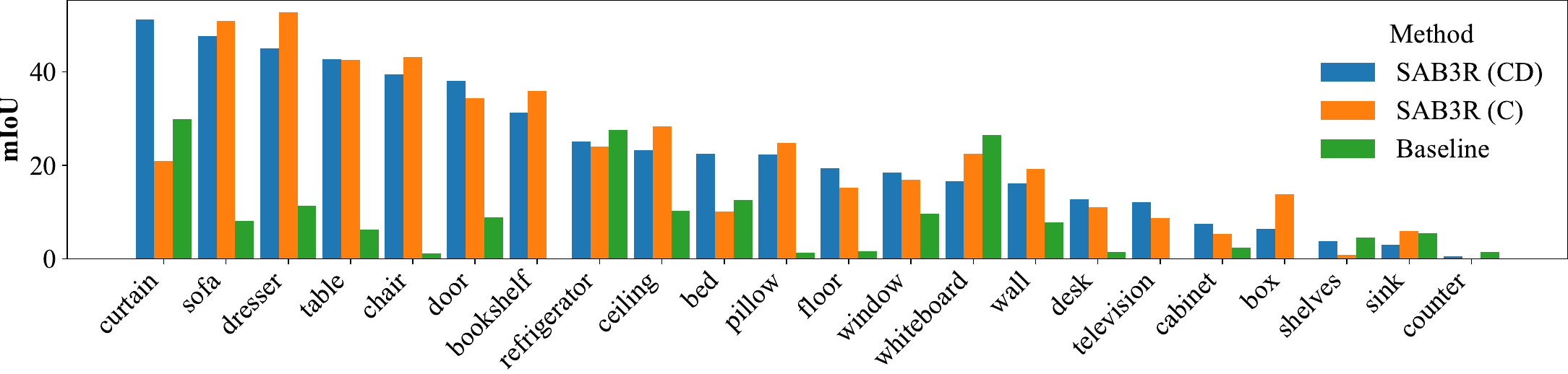}
\caption{\textbf{mIoU Analysis on Frequently Occurring Objects Across Three Methods (Sparse View = 3).} This plot compares mIoU values for frequently appearing objects, illustrating performance differences between our methods and the pipeline approaches and providing insights into the superior results achieved by our methods.}
    \vspace{-5px}
    \label{fig:miou}
\end{figure*}
\vspace{-10pt}
\paragraph{Relative Camera Pose}

Next, we evaluate for the task of relative pose estimation on the CO3Dv2~\cite{reizenstein21co3d} dataset. CO3Dv2 contains 6 million frames extracted from approximately 37k videos, covering 51 MS-COCO categories.

We compare our method’s relative camera pose results with popular approaches like RelPose~\cite{relpose}, RelPose++~\cite{relposepp}, PoseReg and PoseDiff~\cite{posediffusion}, RayDiff~\cite{raydiffusion}, DUSt3R~\cite{dust3r} and MASt3R~\cite{mast3r_arxiv24} in Tab.~\ref{tab:multi-view-pose-regression}. Our experiments show that our method performs comparably to the original MASt3R~\cite{mast3r_arxiv24}, indicating that catastrophic forgetting is not an issue. These results reinforce that \ours retains strong relative camera pose capabilities and can reliably estimate camera poses from unposed images. However, in both 3D tasks, incorporating DINO features does not improve the model's 3D reasoning capabilities.

% \input{figs/tum_vis}
% \input{table/mvs_resconstruction}
% \paragraph{3D Reconstruction}

% We evaluate the quality of our full 3D reconstructions following the global alignment procedure used in DUSt3R's pipeline. This alignment step simplifies the 3D reconstruction task, which is why our reported results for MASt3R differ slightly from those in the original paper. We assess our model's 3D reconstruction performance on the DTU dataset~\cite{jensen2014large} in Table~\ref{tab:mvs_dtu}, where we also align the predictions to the ground-truth coordinate system. We follow the protocol of DUSt3R~\cite{dust3r} and MASt3R~\cite{mast3r_arxiv24} for evaluation.

% While our model after distillation does not reach the accuracy levels of DUSt3R~\cite{dust3r} or MASt3R~\cite{mast3r_arxiv24}, this is likely due to our use of a considerably smaller dataset for fine-tuning \ours, which may have impacted 3D reconstruction performance. To demonstrate that the quality drop remains within acceptable bounds, we include visualizations for reference.
\subsection{Zero-Shot Open Vocabulary Tasks}
\label{subsec:open-vocab}

\paragraph{Zero-Shot Transfer to Semantic Segmentation}

We evaluate the semantic features learned by \ours through zero-shot semantic segmentation on two standard benchmarks: Pascal VOC~\cite{Everingham15} and ADE20K~\cite{zhou2019semantic}. As shown in Table~\ref{tab:zeroshot_combined}, we follow the evaluation protocol of SAM-CLIP~\cite{wang2024samclipmergingvisionfoundation}, with the key distinction that \ours produces dense, pixel-level predictions. Notably, \ours outperforms SAM-CLIP on the more challenging ADE20K dataset, which includes 150 semantic categories. While it does not surpass SAM-CLIP on Pascal VOC, it achieves competitive results and exceeds the performance of the teacher model, FeatUp-upsampled MaskCLIP~\cite{dong2023maskclipmaskedselfdistillationadvances}. We attribute these gains primarily to improved segmentation of large, structurally coherent objects (e.g., curtain, floor, desk). This observation aligns with findings from LeRF~\cite{kerr2023lerflanguageembeddedradiance}, which suggest that models with 3D reasoning capabilities tend to yield stronger semantic segmentation performance. Additional qualitative results, including PCA visualizations of the learned 2D feature space, are included in the supplementary material.

\begin{figure}[t!]
\begin{center}
\includegraphics[width=\linewidth]{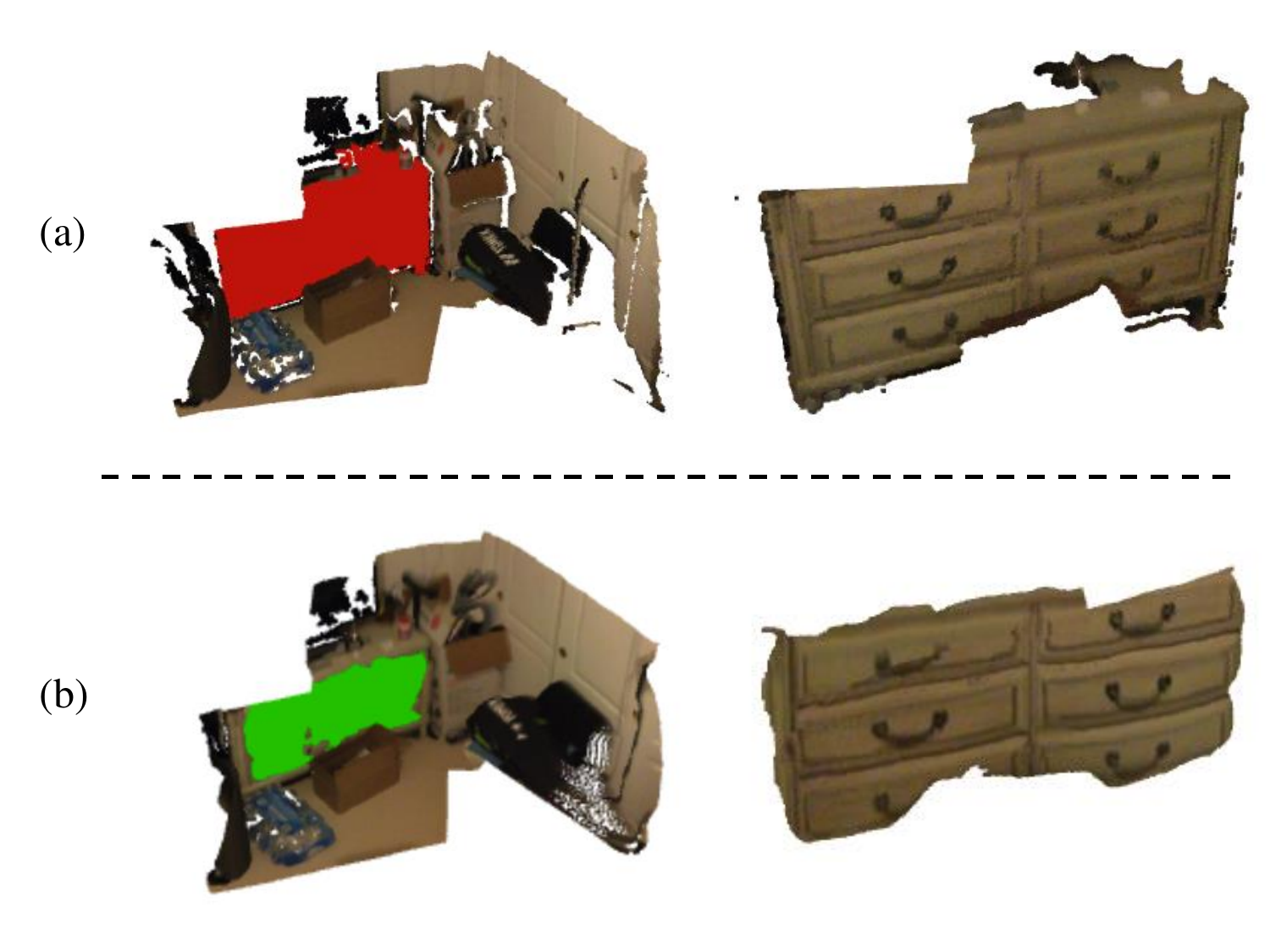}
\vspace{-5px}
\caption{\textbf{Qualitative Example of \textit{Map and Locate}.} This figure illustrates an example from our benchmark. In (a), the ground truth annotation for the scene is highlighted in red, with the dresser segmented from the rest of the scene on the left. In (b), the predictions from \ours are highlighted in green, and the predicted dresser is similarly segmented on the right. These segmented results are subsequently used to compute evaluation metrics.\vspace{-15px}}
\label{fig:quali} 
\vspace{-5px}
\end{center}
\end{figure}
\subsection{A Novel Task - \textit{Map and Locate}}
\label{sec:map_and_locate}

We use MASt3R~\cite{mast3r_arxiv24} and FeatUp~\cite{fu2024featup} as teacher models and adopt them as our primary baselines. Additionally, we report the performance of LSM~\cite{lsm} on this new task for comparison. We present the results in Table~\ref{tab:sparse_view_metrics}. Our method, \ours, consistently outperforms the baseline across all sparse view settings (views = 2, 3, 4) and evaluation metrics, demonstrating strong performance on the \textit{Map and Locate} task. Notably, \ours achieves a 3$\times$ speedup in inference compared to the baseline, as it operates in an end-to-end manner, whereas the baseline relies on a two-stage pipeline involving separate models for reconstruction and segmentation. In terms of semantic quality, measured by mIoU and accuracy, \ours surpasses the baseline by a substantial margin, highlighting its effectiveness in jointly performing 3D reconstruction and open-vocabulary segmentation without pose supervision. For completion metrics, which assess the geometric fidelity of reconstructed semantic objects, \ours also consistently outperforms the baseline under all sparse view configurations. Interestingly, we observe no clear correlation between the number of input views and overall performance. We hypothesize that additional views improve results when they focus on overlapping regions or specific objects, enabling the model to better infer structure and semantics. However, performance may degrade when added views are sparsely distributed across unrelated parts of the scene, leading to reduced overlap and more fragmented supervision during reconstruction.

In Fig.~\ref{fig:miou}, our model demonstrates significant improvements over the baseline in large furniture categories such as sofas, dressers, tables, and chairs. It also successfully recognizes items like bookshelves and televisions, which the baseline fails to detect. Across most categories, our model achieves substantially higher scores, showcasing its strong semantic understanding and superior 3D reconstruction capabilities. Furthermore, it exhibits the ability to identify smaller objects and less common items, underscoring its versatility and robustness. In Fig.~\ref{fig:quali}, we showcase an example of mapping and locating a \textit{dresser} across two images. In part (b) of the qualitative example, the predicted segmentation demonstrates remarkable accuracy compared to the ground truth shown in part (a), highlighting the effectiveness of our model \ours.

LSM~\cite{lsm} demonstrates strong performance when operating on two input views, benefiting from its ability to jointly estimate geometry, semantics, and appearance in a single feed-forward pass. However, extending LSM to more than two views is non-trivial, as its point transformer architecture and Gaussian fusion strategy are designed specifically for dual-view inputs. Moreover, while LSM employs a powerful frozen segmentation backbone that contributes to its accuracy, this comes at the cost of significantly higher computational complexity—both in terms of FLOPs and parameter count—compared to our more lightweight and efficient baseline model \ours.

\vspace*{-5pt}
\section{Conclusion}
\label{sec:conclusion}

Our experiments validate the central insight of this work: 3D open-vocabulary segmentation and 3D reconstruction can be effectively unified through the proposed \textit{Map and Locate} task. Unlike existing approaches that rely on pre-scanned point clouds or posed RGB-D sequences, our formulation accepts unposed video as input—offering a more realistic and scalable setting for embodied agents.

The \textit{Map and Locate} benchmark demonstrates how spatial mapping and semantic understanding can be performed simultaneously, requiring models to reason over both 3D structure and 2D semantics. We present \ours, a simple yet effective baseline that distills 2D foundation models into a unified model capable of predicting 3D point maps along with dense CLIP and DINOv2 features in a single forward pass. Despite its simplicity, \ours performs competitively across both reconstruction and segmentation metrics, while remaining significantly more efficient than multi-stage baselines.

Overall, our findings demonstrate the feasibility of unifying recognition, reconstruction, and reorganization within a single model, offering a more efficient and scalable approach to 3D scene understanding. We hope the \textit{Map and Locate} task serves as a testbed for advancing real-world embodied perception research.

\section{Acknowledgement}
\label{sec:ack}

The authors gratefully acknowledge Research Computing at the University of Virginia for providing the computational resources and technical support that made the results in this work possible. \href{https://rc.virginia.edu}{Research Computing at UVA}.
{
    \small
    \bibliographystyle{ieeenat_fullname}
    \bibliography{main}
}

\clearpage
\appendix
\setcounter{page}{1}
\maketitlesupplementary

In Sec.~\ref{sec:experiments_details}, we provide additional details about the experiments conducted in this work, including a discussion of the software used in \ours and a detailed breakdown of each experiment. Comprehensive analysis and visualizations of our novel task, \textit{Map and Locate}, are provided in Sec.~\ref{sec:find_and_locate}, including both successful and failure cases from our experiments. Sec.~\ref{sec:additional_vis} presents supplementary visualizations of the features generated by CLIP~\cite{radford2021learningtransferablevisualmodels} and DINOv2~\cite{oquab2024dinov2learningrobustvisual}. Finally, we discuss the limitations of our approach in Sec.~\ref{sec:limitations}.

\section{More Experiment Details}
\label{sec:experiments_details}
\subsection{Teacher Models and Frameworks}

\paragraph{CLIP \& MaskCLIP}
Vision and language models are trained to generate aligned feature embeddings using a contrastive objective.
The original CLIP family of models was proposed by \citet{radford2021learningtransferablevisualmodels} and included a wide variety of architectures in a private dataset of 400M image-text pairs called WIT.  More recently, \citet{openclip} trained several CLIP models using several architectures trained on publicly available datasets. In \ours, we used MaskCLIP~\cite{dong2023maskclipmaskedselfdistillationadvances}, which enhances CLIP pretraining by introducing masked self-distillation. This transfers knowledge from full-image representations to masked-image predictions. This approach complements the vision-language contrastive objective by focusing on local patch representations while aligning features with indirect supervision from language. Additionally, MaskCLIP incorporates local semantic supervision into the text branch, further improving pretraining performance. We follow suggestions from FeatUp~\cite{fu2024featup} that MaskCLIP~\cite{dong2023maskclipmaskedselfdistillationadvances} has better local semantic feature compare with CLIP~\cite{radford2021learningtransferablevisualmodels}.

\paragraph{MASt3R} 

MASt3R~\cite{mast3r_arxiv24} was trained on an extensive multi-view dataset comprising 5.3 million real-world image pairs and 1.8 million synthetic pairs. The real-world data includes diverse scenarios from ARKitScenes~\cite{dehghan2021arkitscenes}, MegaDepth~\cite{MDLi18}, 3DStreetView~\cite{Zamir_2016}, and IndoorVL~\cite{lee2021largescalelocalizationdatasetscrowded}. The synthetic data was generated using the Habitat simulator~\cite{savva2019habitatplatformembodiedai}, covering indoor, outdoor, and landmark environments.

Our model is finetuned on top of MASt3R, leveraging Habitat-Sim~\cite{savva2019habitatplatformembodiedai}, ScanNet++\cite{yeshwanth2023scannethighfidelitydataset3d}, and Co3Dv2\cite{reizenstein21co3d}, ARKitScenes~\cite{dehghan2021arkitscenes} and BlenderMVS~\cite{yao2020blendedmvs}.

\paragraph{FeatUp}

FeatUp~\cite{fu2024featupmodelagnosticframeworkfeatures} is a framework designed to enhance spatial resolution in deep features for tasks like segmentation and depth prediction. It addresses the loss of spatial detail caused by pooling in traditional networks using two approaches: guided upsampling with high-resolution signals in a single pass and reconstructing features at arbitrary resolutions with an implicit model. Both methods use a multi-view consistency loss inspired by NeRFs to maintain feature semantics.

FeatUp integrates seamlessly into existing pipelines, boosting resolution and performance without re-training. Experiments demonstrate its superiority over other methods in tasks such as segmentation, depth prediction, and class activation map generation. In \ours, we find the MaskCLIP variant of FeatUp model can also perform zero-shot semantic segmentation and we use it as our teacher model for distillation.

\begin{table}[H]
\centering
\caption{\textbf{Checkpoint Details.} Information about the pre-trained checkpoints used in this work, including source and license.}
\label{tab:checkpoint_details}
\begin{tabular}{|l|l|l|}
\hline
\textbf{Checkpoint}                  & \textbf{Source Link}                                                                                                                   & \textbf{License}       \\ \hline
FeatUp MaskCLIP               & \href{https://marhamilresearch4.blob.core.windows.net/feature-upsampling-public/pretrained/no_norm/maskclip_jbu_stack_cocostuff.ckpt}{MaskCLIP} & MIT                   \\ \hline
MASt3R        & \href{https://download.europe.naverlabs.com/ComputerVision/MASt3R/MASt3R_ViTLarge_BaseDecoder_512_catmlpdpt_metric.pth}{MASt3R} & CC BY-NC-SA 4.0       \\ \hline
\end{tabular}
\end{table}

We list the checkpoints used in \ours in Tab.~\ref{tab:checkpoint_details}, detailing the FeatUp MaskCLIP variant and MASt3R, along with their source links and license information.

% \subsection{Hardware}

% Each checkpoint is optimized around 3 days, using either 8 A40 GPUs or 4 A100 80GB GPUs.

% \subsection{Training Details}

% \paragraph{Dense 2D Feature Head}

% \paragraph{Hyperparameter}

% \paragraph{Multi-Task Distillation}

% \subsection{Loss Coefficients}

\subsection{Experiments Details}

\paragraph{Monocular Depth}

In the main text, we benchmark \ours on the outdoor dataset KITTI~\cite{Geiger2013IJRR} and the indoor dataset NYUv2~\cite{Silberman:ECCV12}. Here, we provide a detailed discussion of the evaluation metrics. Following DUSt3R, we use two commonly adopted metrics in monocular depth estimation:

\begin{itemize}
    \item Absolute Relative Error (AbsRel): This measures the relative error between the ground truth depth $y$ and the predicted depth $\hat{y}$, defined as:
    \[
    \text{AbsRel} = \frac{|y - \hat{y}|}{y}.
    \]

    \item Prediction Threshold ($\delta_{1.25}$): This evaluates the fraction of predictions within a given threshold and is defined as:
    \[
    \delta_{1.25} = \frac{\max\left(\frac{\hat{y}}{y}, \frac{y}{\hat{y}}\right) < 1.25}{\text{Total Predictions}}.
    \]
\end{itemize}

These metrics allow for comprehensive evaluation of depth prediction accuracy and robustness across different datasets.
\begin{figure}[t!]
    \centering
    \includegraphics[width=0.7\linewidth]{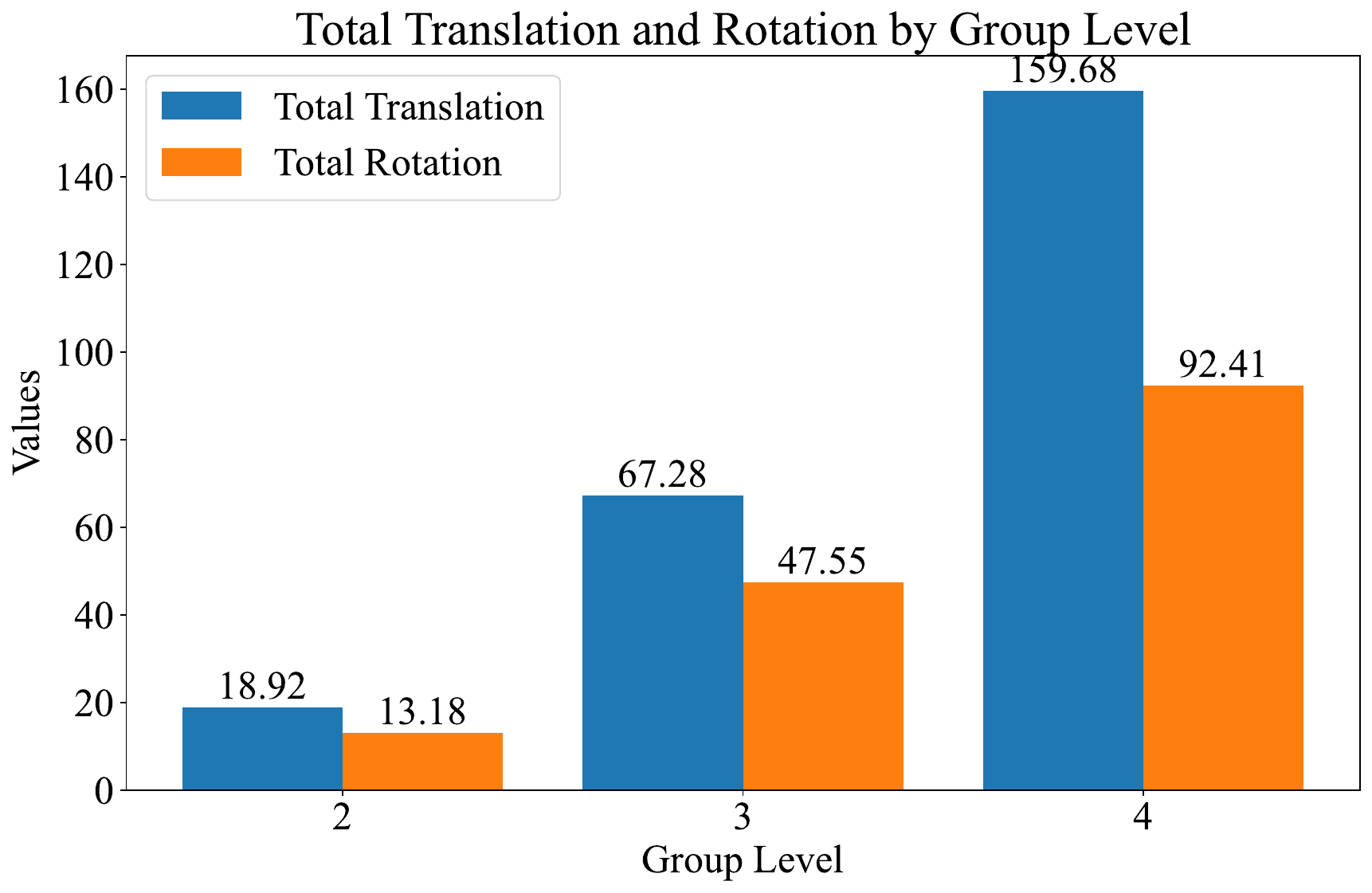}
    \caption{\textbf{Camera Distributions.} Camera translation differences and rotation differences at different group levels.}
    \label{fig:camera-vis}
\end{figure}
\paragraph{Relative Camera Pose}

We evaluate \ours on the task of relative pose estimation using the CO3Dv2 dataset~\cite{reizenstein21co3d}. To assess the relative pose error for each image pair, we report the Relative Rotation Accuracy (RRA) and Relative Translation Accuracy (RTA). For evaluation, we select a threshold $\tau = 15^\circ$ and report $\text{RRA@15}$ and $\text{RTA@15}$, representing the percentage of image pairs where the errors in rotation and translation are below the threshold $\tau$.

The rotation error $\text{e}_\text{rot}$ and translation error $\text{e}_\text{trans}$ for each image pair are computed as:
\[
\text{e}_\text{rot} = \arccos\left(\frac{\text{trace}(\mathbf{R}^\top \hat{\mathbf{R}}) - 1}{2}\right),
\]
\[
\text{e}_\text{trans} = \arccos\left(\frac{\mathbf{t}^\top \hat{\mathbf{t}}}{\|\mathbf{t}\| \|\hat{\mathbf{t}}\|}\right),
\]
where $\mathbf{R}$ and $\hat{\mathbf{R}}$ are the ground truth and predicted rotation matrices, and $\mathbf{t}$ and $\hat{\mathbf{t}}$ are the ground truth and predicted translation vectors.

We also report the mean Average Accuracy (mAA@30), defined as the area under the accuracy curve of the angular differences for $\min(\text{RRA@30}, \text{RTA@30})$. The mAA@30 is calculated as:
\[
\text{mAA@30} = \frac{1}{30} \int_0^{30} \min(\text{RRA@}\theta, \text{RTA@}\theta) \, d\theta,
\]
where $\theta$ represents the threshold angle in degrees.

\paragraph{Zero-Shot Semantic Segmentation}

For zero-shot semantic segmentation, we largely follow the approach outlined by Ranasinghe et al.\cite{PERCEPTUAL}, utilizing 80 prompt templates introduced by Radford et al~.\cite{radford2021learningtransferablevisualmodels, wang2024samclipmergingvisionfoundation}. Class names are embedded into these prompts, and text embeddings are generated using the text encoder. We then compute the cosine similarity between each text embedding and the corresponding pixel feature—extracted directly from the CLIP head. The class with the highest cosine similarity is assigned as the predicted class for each pixel.

The class predictions are subsequently resized to match the original image dimensions, and the mean Intersection over Union (mIoU) is computed for evaluation. Unlike prior methods, our approach eliminates the concept of patches. Instead, because the CLIP head directly generates per-pixel features, we can seamlessly perform top-1 matching between semantic classes and pixel features, bypassing the need for patch-based processing.

\section{Additional \textit{Map and Locate} Details}
\label{sec:find_and_locate}
\subsection{Dataset Summary}

We evaluate our \textit{Map and Locate} framework using the ScanNet dataset~\cite{dai2017scannet}, a large-scale indoor scene dataset that provides RGB-D sequences, camera poses, and both semantic and instance-level annotations. Specifically, we select 24 scenes from the validation split, each containing diverse object layouts and camera trajectories. Across these scenes, there are a total of 2,261 annotated objects with semantic and instance-level ground truth.

For evaluation, we construct 2 sets of image groups for each scene, where each group comprises 2, 3, or 4 images. The image selection follows two principles:
\begin{itemize}
\item \textbf{Object visibility:} Objects in each group are visible across multiple images to ensure reliable localization and mapping.
\item \textbf{Viewpoint diversity:} Selected images capture varying camera viewpoints to test robustness to occlusion and perspective changes.
\end{itemize}

In total, this results in 144 image groups (2 sets per scene $\times$ 24 scenes $\times$ 3 group sizes). Each group is paired with its corresponding RGB images, depth maps, camera intrinsics and extrinsics, as well as semantic and instance segmentation labels. This setup provides a comprehensive benchmark for evaluating both mapping accuracy and object localization performance under realistic and challenging scene configurations.

\subsection{Dataset Visualizations}

We present a dataset statistics visualization in Fig.~\ref{fig:camera-vis}, showing camera translation differences and rotation differences. Translation differences are computed as the Euclidean distance between translation vectors, \( d_{\text{translation}} = \|\mathbf{t}_1 - \mathbf{t}_2\|_2 \), and rotation differences are calculated as the geodesic distance on \( SO(3) \), \( d_{\text{rotation}} = \|\mathbf{r}_{\Delta}\|_2 \), where \( \mathbf{r}_{\Delta} \) is the axis-angle representation of the relative rotation \( \mathbf{R}_{\Delta} = \mathbf{R}_1^{-1} \mathbf{R}_2 \). These metrics highlight the variability in camera poses across the dataset. We observe that as the number of views increases, both camera translation differences and rotation differences grow. Despite this, our results demonstrate consistent performance across all group levels, highlighting the robustness of our algorithm.
\begin{figure*}[t!]
    \centering
    \includegraphics[width=\linewidth]{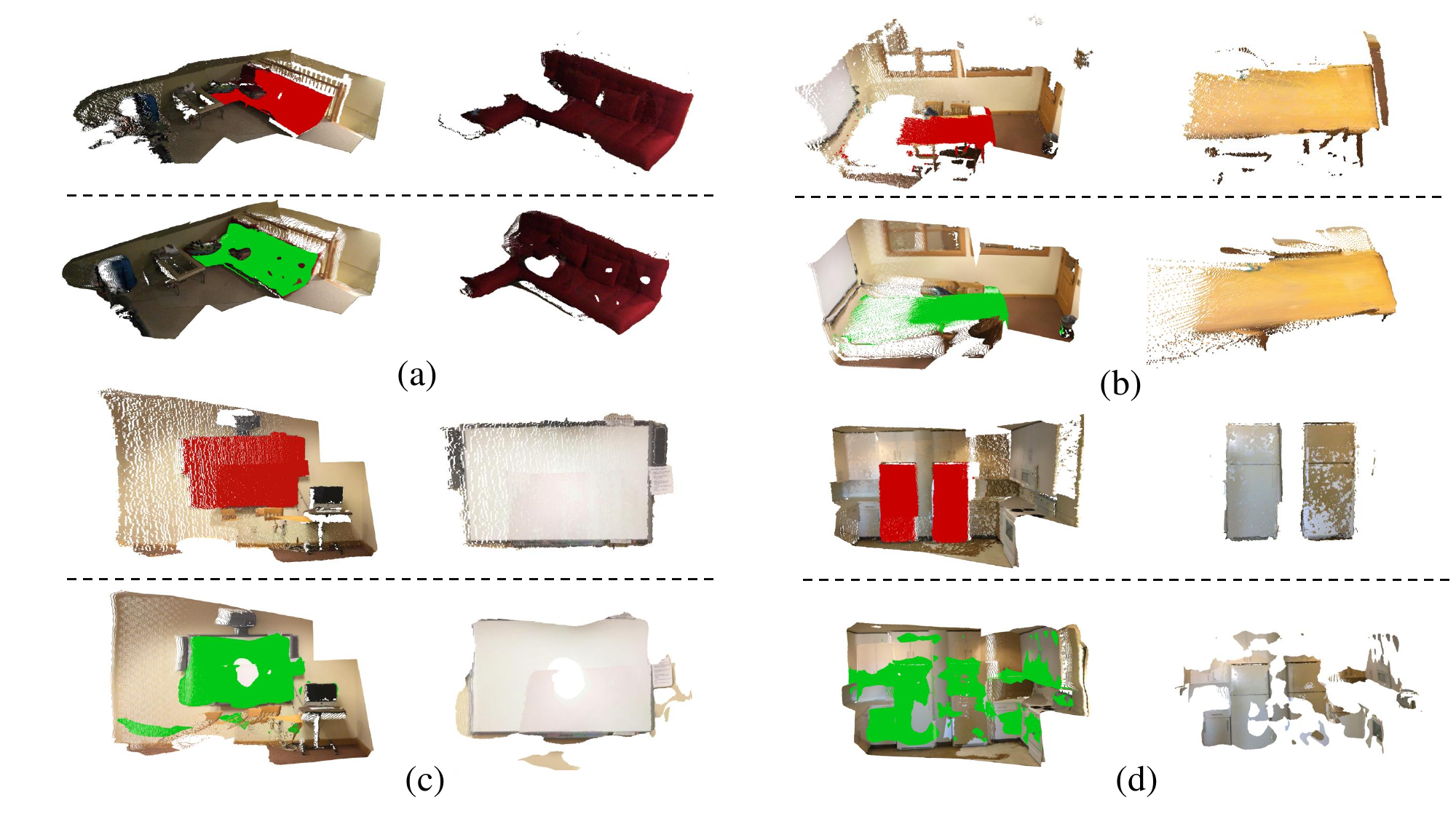}
    \caption{\textbf{Qualitative Examples of \textit{Map and Locate} with \ours.} Panels (a), (b), and (c) illustrate successful examples of 3D scene reconstruction and accurate object segmentation. In each sub-group, the top row shows the ground truth, with the target objects highlighted in red, accompanied by visualizations of segmented objects for each ground truth target. The bottom row presents the predicted results, where the segmented objects are shown in green, with the extracted objects displayed on the right for clarity. Panel (d) provides an example of a failure case.
    }
    \label{fig:qualitative_map_and_local_additional}
\end{figure*}

\subsection{More Qualitative Examples}

Fig.~\ref{fig:qualitative_map_and_local_additional} presents additional qualitative examples demonstrating the performance of \textit{Map and Locate} with \ours.

\section{Additional visualization}
\label{sec:additional_vis}

\begin{figure*}[t!]
    \centering
    \includegraphics[width=0.9\linewidth]{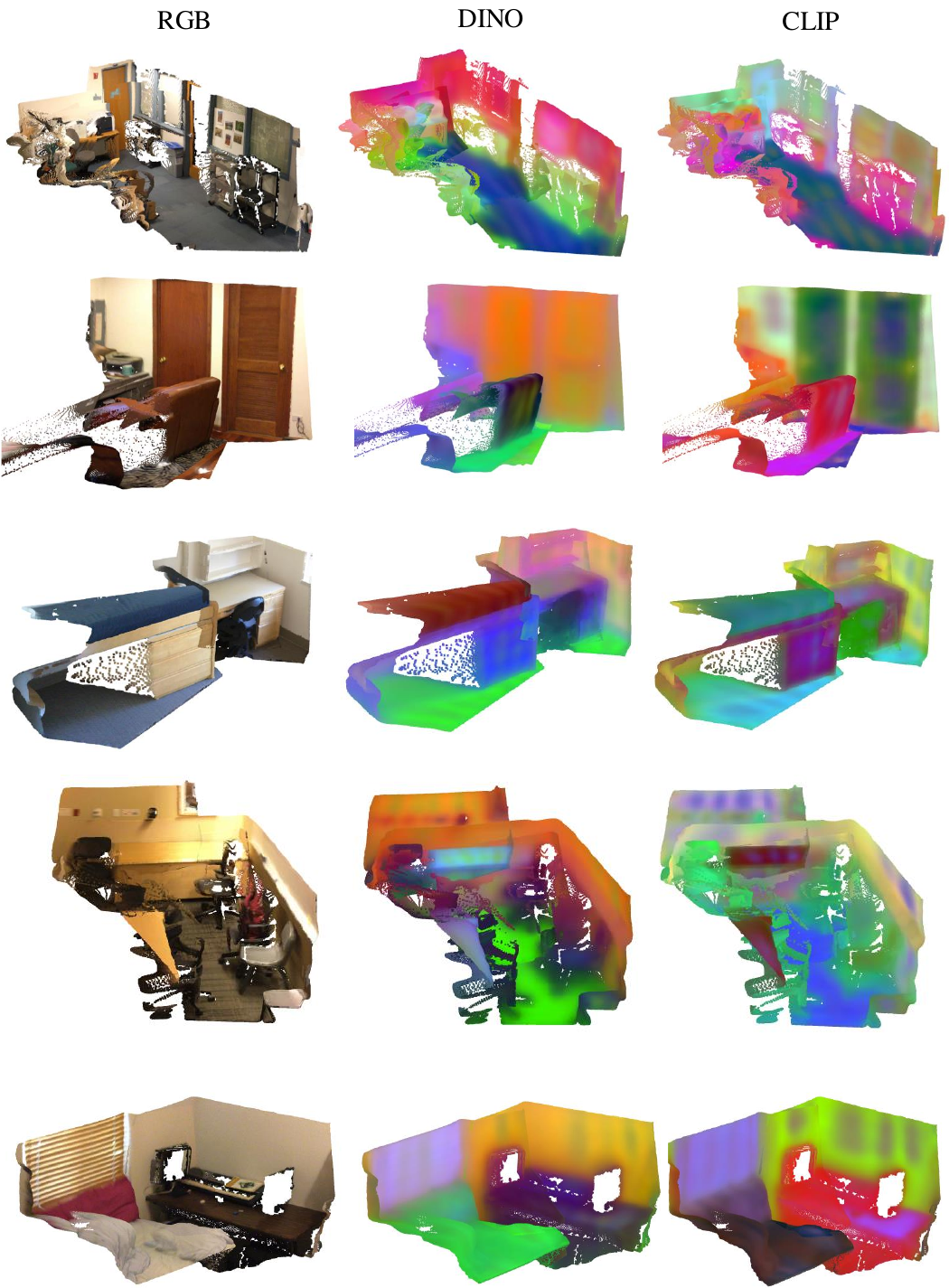}
    \caption{\textbf{3D Feature Visualizations.} Additional visualizations of 3D features are presented for DINO and CLIP, alongside the original RGB 3D point map for reference.}
    \label{fig:additional_feature_vis}
\end{figure*}

Fig.~\ref{fig:additional_feature_vis} presents additional visualizations of 3D features from DINO~\cite{oquab2024dinov2learningrobustvisual} and CLIP~\cite{radford2021learningtransferablevisualmodels}. The visualizations highlight distinct features for different objects. Predicted RGB is provided as a reference.
 
\section{Limitations}
\label{sec:limitations}

Our study is constrained by limited computational resources, which restricted us from training the model for more epochs, potentially resulting in under-trained checkpoints. Additionally, predicting dense features significantly increases vRAM requirements, further limiting our ability to optimize the model fully. Due to these resource constraints, we were unable to use the entire pre-training dataset for fine-tuning, which may have prevented the model from achieving its best possible performance. Our novel task, \textit{Map and Locate}, relies on the ScanNet dataset, which, despite its comprehensiveness, is primarily biased toward indoor environments. Extending this work to more diverse datasets, including outdoor or dynamic scenes, represents an interesting direction for future works.

\end{document}